\pdfoutput=1

\documentclass[11pt]{article}

\usepackage[]{acl}

\usepackage{times}
\usepackage{latexsym}

\usepackage[T1]{fontenc}

\usepackage[utf8]{inputenc}

\usepackage{microtype}

\usepackage{booktabs}
\usepackage{multirow}
\usepackage{framed}
\usepackage{xcolor}
\usepackage{graphicx}
\usepackage{bm}
\usepackage{rotating}
\usepackage{listings}
\usepackage{graphics}
\usepackage{float} 
\usepackage{subfig} 
\usepackage{multicol}
\usepackage{amsmath}
\usepackage{amssymb}
\usepackage{pifont}

\newcommand{\tabincell}[2]{\begin{tabular}{@{}#1@{}}#2\end{tabular}}

\title{Towards Emotional Support Dialog Systems}

\author{\textbf{Siyang Liu${^{1,2}}$\thanks{\ \ Equal Contribution.
} , Chujie Zheng$^1$\footnotemark[1] , Orianna Demasi$^3$, Sahand Sabour$^1$, Yu Li$^3$, } \\
\textbf{Zhou Yu$^4$, Yong Jiang$^2$, Minlie Huang$^1$\thanks{\ \ Corresponding author.}} \\
  \small $^1$The CoAI group, DCST, Institute for Artificial Intelligence, State Key Lab of Intelligent Technology and Systems, \\
  \small $^1$Beijing National Research Center for Information Science and Technology, Tsinghua University, Beijing 100084, China \\
  \small $^2$Tsinghua-Berkeley Shenzhen Institute, Tsinghua Shenzhen International Graduate School, \\ 
  \small $^2$Tsinghua University, Shenzhen, China \\
  \small $^3$University of California, Davis \quad $^4$Columbia University \\
  {\small \tt siyang-l18@mails.tsinghua.edu.cn, chujiezhengchn@gmail.com, aihuang@tsinghua.edu.cn} \\
}

\date{}

\begin{document}
\maketitle

\begin{abstract}

Emotional support is a crucial ability for many conversation scenarios, including social interactions, mental health support, and customer service chats. Following reasonable procedures and using various support skills can help to effectively provide support.
However, due to the lack of a well-designed task and corpora of effective emotional support conversations, research on building emotional support into dialog systems remains untouched.
In this paper, we define the Emotional Support Conversation (ESC) task and propose an ESC Framework, which is grounded on the Helping Skills Theory \cite{hill2009helping}.
We construct an Emotion Support Conversation dataset (ESConv) with rich annotation (especially support strategy) in a help-seeker and supporter mode.
To ensure a corpus of high-quality conversations that provide examples of effective emotional support, we take extensive effort to design training tutorials for supporters and several mechanisms for quality control during data collection.
Finally, we evaluate state-of-the-art dialog models with respect to the ability to provide emotional support.
Our results show the importance of support strategies in providing effective emotional support and the utility of ESConv in training more emotional support systems \footnote{Our data and codes are available at\\\url{https://github.com/thu-coai/Emotional-Support-Conversation}.}.

\end{abstract}

\section{Introduction}

Emotional support (ES) aims at reducing individuals' emotional distress and helping them understand and work through the challenges that they face \cite{burleson2003emotional,langford1997social,heaney2008social}.
It is a critical capacity to train into dialog systems that interact with users on daily basis \cite{van2012bdi, zhou2020design}, particularly for settings that include social interactions (accompanying and cheering up the user), mental health support (comforting a frustrated help-seeker and helping identify the problem), customer service chats (appeasing an angry customer and providing solutions), etc.
Recent research has also shown that people prefer dialog systems that can provide more supportive responses \cite{rains2020support}.

\begin{figure}[t]
  \centering
  \includegraphics[width=\linewidth]{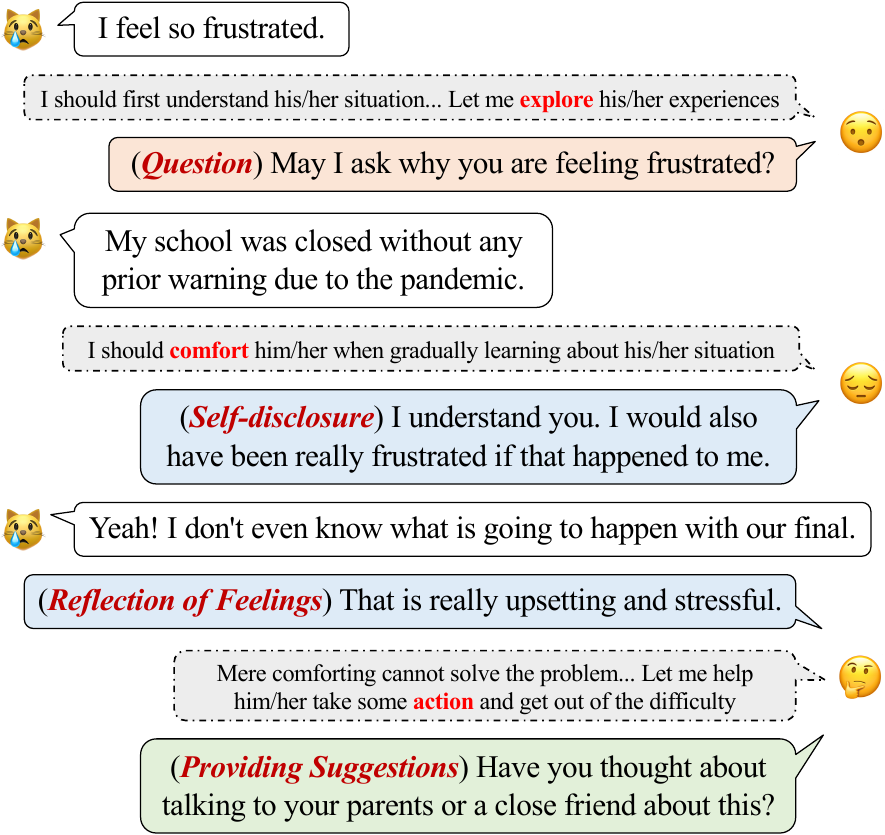}
  \caption{
    An example chat showing effective emotional support (adapted from ESConv) being provided to the help-seeker(left) by the supporter(right).
    The \textcolor[RGB]{190,37,14}{\textit{\textbf{support strategies (skills)}}} used by the supporter are marked in the parentheses before the utterances.
    The {\color{red} \textbf{red bold texts}} in the dashed boxes highlight the three stages of our proposed ESC Framework (Figure \ref{fig:framework}).
  }
  \label{fig:intro_example}
  \vspace{-2mm}
\end{figure}

Research has shown that providing emotional support is not intuitive \cite{burleson2003emotional}, so procedures and conversational skills have been suggested \cite{hill2009helping} to help provide better support through conversation. Such skills can be seen in the example conversation that we collected and is shown in Figure \ref{fig:intro_example}. 
To identify the causes of the help-seeker's distress, the supporter first explores the help-seeker's problems.
Without exploration, the support is unlikely to understand the help-seeker's experiences and feelings, and thus it may be \textit{offensive} or even \textit{harmful} if the supporter would give irrelevant advice, like `\textit{You could go for a walk to relax}'.
While learning about the help-seeker's situation, the supporter may express understanding and empathy to relieve the help-seeker's frustration by using various skills (e.g., \textit{Self-disclosure}, \textit{Reflection of Feelings}, etc.).
After understanding the help-seeker's problem, the supporter may offer suggestions to help the help-seeker cope with the problem.
If the supporter only comforts the help-seeker without any inspiration for action to change, the supporter may not effectively help the help-seeker's emotions improve.
Finally, during the data collection of this example conversation, the help-seeker reported that their emotion intensity decreased from 5 to 2 (emotion intensity is labeled in our corpus, we give detailed annotations of this conversation example in Appendix \ref{app:example}), which indicates the effectiveness of the ES provided by the supporter.

\begin{figure}[t]
  \centering
  \includegraphics[width=0.75\linewidth]{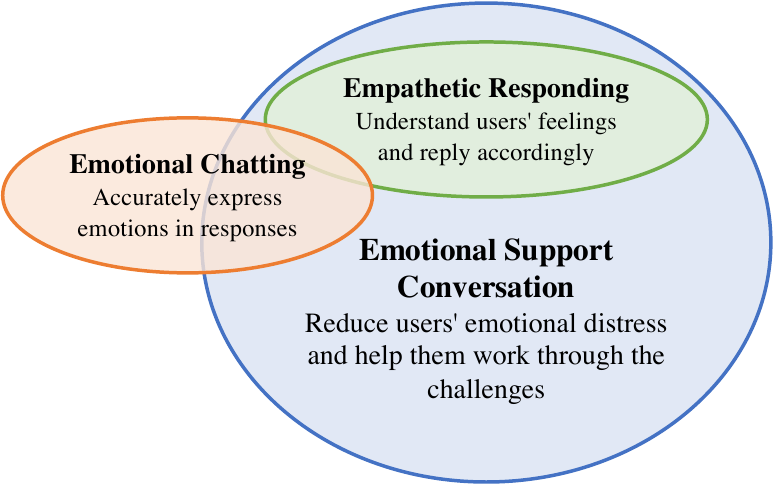}
  \caption{
  Emotional support conversations (our work) can include elements of emotional chatting \cite{zhou2018emotional} and empathetic responding\cite{rashkin-etal-2019-towards}.
  }
  \label{fig:relation}%
  \vspace{-2mm}
\end{figure}

Despite the importance and complexity of ES, research on data-driven ES dialog systems is limited due to a lack of both task design and relevant corpora of conversations that demonstrate diverse ES skills in use.
{First}, existing research systems that relate to emotional chatting \cite{zhou2018emotional} or empathetic responding \cite{rashkin-etal-2019-towards} return messages that are examples of emotion or empathy and are thus limited in functionality, as they are not capable of many other skills that are often used to provide effective ES \cite{hill2009helping}.
Figure \ref{fig:relation} illustrates the relationship between the three tasks and we provide further discussion in Section \ref{subsec:comparison}.
{Second}, people are not naturally good at being supportive, so guidelines have been developed to train humans how to be more supportive.
Without trained individuals, existing online conversation datasets\cite{sharma-etal-2020-computational, rashkin-etal-2019-towards, zhong-etal-2020-towards, sun-etal-2021-psyqa} do not naturally exhibit examples or elements of supportive conversations.
As a result, data-driven models that leverage such corpora \cite{radford2019language, zhang-etal-2020-dialogpt, roller2020recipes} are limited in their ability to explicitly learn how to utilize support skills and thus provide effective ES.

In this paper, we define the task of Emotional Support Conversation (\textbf{ESC}), aiming to provide support through social interactions (like the interactions between peers, friends, or families) rather than professional counseling, and propose an \textbf{ESC Framework}, which is grounded on the Helping Skills Theory \cite{hill2009helping} and tailored to be appropriate for a dialog system setting (Figure \ref{fig:framework}).
We carefully design the ESC Framework for a dialog system setting by adapting relevant components of Hill's Helping Skills model of conversational support.  
The ESC Framework proposes three stages (\textit{Exploration}, \textit{Comforting} and \textit{Action}), where each stage contains several support strategies (or skills). 
To facilitate the research of emotional support conversation, we then construct an Emotional Support Conversation dataset, \textbf{ESConv}, and take multiple efforts to ensure rich annotation and that all conversations are quality examples for this particularly complex dialog task.
ESConv is collected with crowdworkers chatting in help-seeker and supporter roles. We design tutorials based on the ESC framework and train all the supporters and devise multiple manual and automatic mechanisms to ensure effectiveness of emotional support in conversations.
Finally, we evaluate the state-of-the-art models and observe significant improvement in the emotional support provided when various support strategies are utilized.
Further analysis of the interactive evaluation results shows the Joint model can mimic human supporters' behaviors in strategy utilization.
We believe our work will facilitate research on more data-driven approaches to build dialog systems capable of providing effective emotional support.

\section{Related Work}
\label{sec:related}

\subsection{Emotional \& Empathetic Conversation}
\label{subsec:comparison}
Figure \ref{fig:relation} intuitively shows the relationships among ESC, emotional conversation, and empathetic conversation.
Emotion has been shown to be important for building more engaging dialog systems \cite{zhou2018emotional, li-etal-2017-dailydialog, zhou-wang-2018-mojitalk, huber2018emotional, huang2020challenges}.
As a notable work of emotional conversation, \citet{zhou2018emotional} propose Emotional Chatting Machine (ECM) to generate emotional responses given a pre-specified emotion.
This task is required to accurately express (designated or not) emotions in generated responses.
While ES may include expressing emotions, such as happiness or sadness, it has a broader aim of reducing the user's emotional distress through the utilization of proper support skills, which is fundamentally different from emotional chatting.
Emotional chatting is merely a basic quality of dialog systems, while ES is a more high-level and complex ability that dialog systems are expected to be equipped with.
Another related task is empathetic responding \cite{rashkin-etal-2019-towards, lin-etal-2019-moel, majumder-etal-2020-mime, zandie2020emptransfo, sharma-etal-2020-computational, zhong-etal-2020-towards, zheng-etal-2021-comae}, which aims at understanding users' feelings and then replying accordingly.
For instance, \citet{rashkin-etal-2019-towards} argued that dialog models can generate more empathetic responses by recognizing the interlocutor's feelings.
Effective ES naturally requires expressing empathy according to the help-seeker's experiences and feelings, as shown in our proposed Emotional Support Framework (Section \ref{subsec:framework}, Figure \ref{fig:framework}).
Hence, empathetic responding is only one of the necessary components of emotional support.
In addition to empathetic responding, an emotional support conversation needs to explore the users' problems and help them cope with difficulty.

\vspace{-1mm}
\subsection{Related Datasets for Emotional Support}
\vspace{-1mm}

Various works have considered conversations of emotional support in a social context, such as on social media or online forums \cite{medeiros2018using, sharma2020computational, hosseini2021takes}.
\citet{medeiros2018using} collected stress-related posts and response pairs from Twitter and classified replies into supportive categories. 
In \cite{sharma2020computational}, the post-response pairs from TalkLife and mental health subreddits are annotated with the communication mechanisms of text-based empathy expression (only the data of the Reddit part is publicly available).
\citet{hosseini2021takes} also collected such post-response pairs from online support groups, which have been annotated as needing or expressing support. 
The dialogues in these corpora are either single-turn interactions (post-response pair) or very short conversations, which limits the potential for effective ES, as ES often requires many turns of interaction \cite{hill2009helping}.

\vspace{-1mm}
\subsection{Emotional Support Dialog Systems}
\vspace{-1mm}

Some traditional dialog systems have applied human-crafted rules to provide emotional support responses \cite{van2012bdi, van2012conversation}. 
A recent system has considered a rule-based algorithm that determines the supportive act used in the response and then selects proper replies from the pre-defined list of candidates \cite{medeiros2018using}. 
Another conversational system designed to provide support for coping with COVID-19 was implemented by identifying topics that users mentioned and then responding with a reflection from a template or a message from a pre-defined lexicon \citep{welch-etal-2020-expressive}.
Few studies have focused on generating supportive responses, and those that have have been limited in scope. 
For example, \citet{shen-etal-2020-counseling} explored how to generate supportive responses via reflecting on user input.

\begin{figure*}[t]
  \centering
  \includegraphics[width=\linewidth]{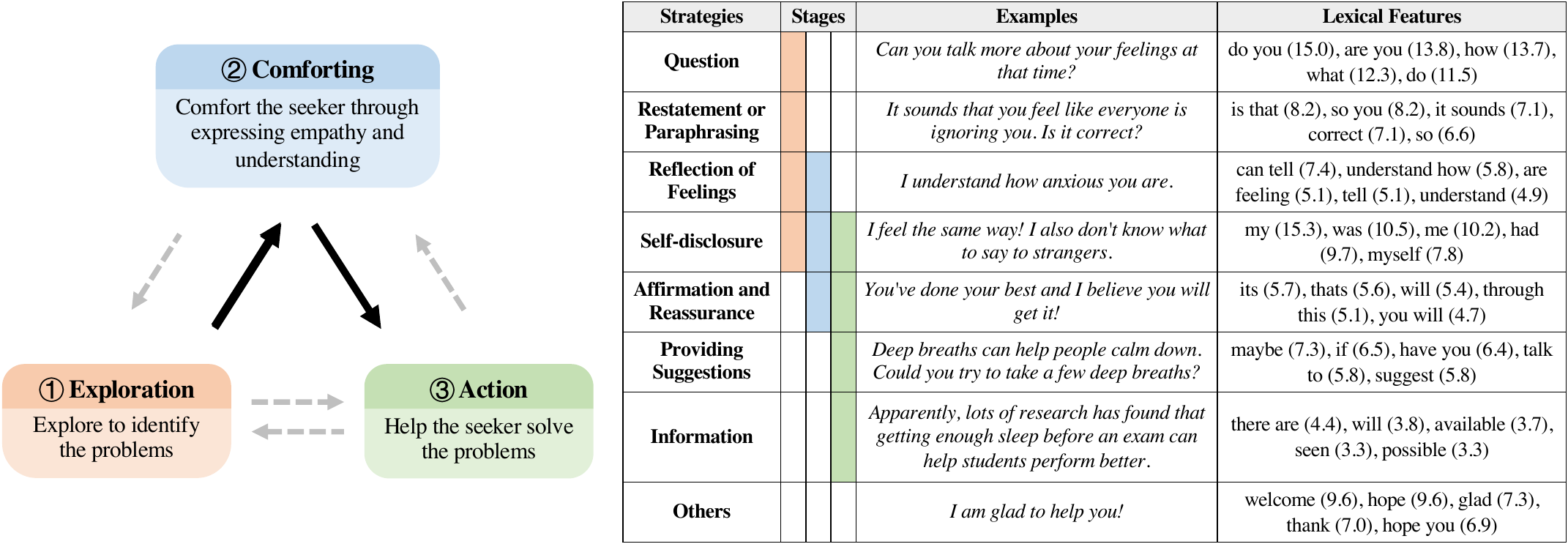}
  \caption{
    Overview of our proposed ESC Framework. It contains three stages and suggested support strategies.
    The procedure of emotional support generally follows the order: \textcircled{1}\textit{Exploration} $\to$ \textcircled{2}\textit{Comforting} $\to$ \textcircled{3}\textit{Action} (as indicated by the \textbf{black} arrows), but it can also be adapted to the individual conversation as needed (indicated by the {\color{gray} \textbf{dashed gray}} arrows).
    The column of ``Lexical Features'' displays top 5 unigrams or bigrams associated with messages that use each strategy in our dataset. Each feature is ranked by the rounded $z$-scored log odds ratios \cite{monroe2008fightin} in the parentheses.
  }
  \label{fig:framework}
  \vspace{-2mm}
\end{figure*}

\vspace{-1mm}
\section{Emotional Support Conversation}
\vspace{-1mm}

\subsection{Task Definition}
\vspace{-1mm}
\label{subsec:task}

When a user is in a bad emotional state, perhaps due to a particular problem, they may seek help to improve their emotional state. 
In this setting, the user can be tagged with a negative emotion label $e$, a emotion intensity level $l$ (e.g., ranging from 1 to 5), and an underlying challenge that the user is going through. 
The supporter (or the system) needs to comfort the user in a conversation with support skills to lower their intensity level. 
Note that the user's state is unknown to the supporter prior to the conversation.
During the conversation, the supporter needs to identify the problem that the user is facing, comfort the user, and then provide some suggestions or information to help the user take action to cope with their problem. 
An emotional support conversation is effective if the intensity level of the user is lowered at the end of the conversation, or more concretely, if the supporter can effectively identify the problem, comfort the user, and provide solutions or suggestions.  

The ESC task has several sub-problems:
(1) Support strategy selection and strategy-constrained response generation.
As shown in our later experiments (Section \ref{subsec:interactive}), the timing of applying strategies is relevant to the effectiveness of ES.
It is thus important that a generated response conforms to a specified strategy.
(2) Emotion state modeling. It is important to model and track the user's emotion state dynamically, both for dynamic strategy selection and for measuring the effectiveness of ESC.
(3) Evaluation of support effectiveness. In addition to the traditional dimension of evaluating a conversation's relevance, coherence, and user engagement, ESC raises a new dimension of evaluating the effectiveness of ES.

\vspace{-1mm}
\subsection{ESC Framework}
\vspace{-1mm}
\label{subsec:framework}

We present an ESC Framework, which characterizes the procedure of emotional support into three stages, each with several suggested support strategies.
We ground the ESC Framework on Hill's Helping Skills Theory \cite{hill2009helping} and adapt it more appropriate for a dialog system setting, aiming to provide support through social interactions (like the interactions between peers, friends, or families) rather than merely professional counseling.
An overview of the conversational stages and strategies in the ESC Framework is shown in Figure \ref{fig:framework}.

\noindent \textbf{Stages} \quad
\citet{hill2009helping} proposes three stages of supporting people: 
{\textit{exploration}} (exploring to help the help-seeker identify the problems), 
{\textit{insight}} (helping the help-seeker move to new depths of self-understanding),
and {\textit{action}} (helping the help-seeker make decisions on actions to cope with the problems).
However, we note that {\textit{insight}} usually requires re-interpreting users' behaviors and feelings, which is both difficult and risky for the supporters without sufficient support experience.
We thus adapt {\textit{insight}} to {\textit{comforting}} (defined as providing support through empathy and understanding).
While it is suggested that emotional support conversations target these three ordered stages, in practice conversations cannot follow a fixed or linear order and must adapt appropriately.
As suggested in \cite{hill2009helping}, the three stages can be flexibly adjusted to meet the help-seeker's needs.

\noindent \textbf{Strategies} \quad
\citet{hill2009helping} also provides several recommended conversational skills for each stage. Some of the described skills are not appropriate\footnote{
For instance, one skill named \textit{challenging} refers to pointing out the discrepancies or irrational beliefs that the help-seeker is unaware of or unwilling to change.
Such skills usually require professional experience, which is too difficult for an average person.
} in a dialog system setting without professional supervision and experience. To adapt these skills appropriate to the dialog system setting, we extract seven methods from these skills (along with an ``Others'' one), which we called \textit{strategies} in our task and hereafter.
We provide a detailed definition of each strategy in Appendix \ref{sec:definition}.

\vspace{-1mm}
\section{Data Collection}
\vspace{-1mm}
\label{sec:collection}

To facilitate the research of emotional support skills in dialog systems, we introduce an Emotional Support Conversation Dataset, \textbf{ESConv}, which is collected in a help-seeker and supporter mode with crowdworkers.
As high-quality conversation examples are needed for this complex task, we took tremendous effort to try to ensure the effectiveness of ES in conversations. Our efforts included the following major aspects:
(1) Because providing conversational support is a skill that must be trained for supporters to be effective \cite{burleson2003emotional}, we design a tutorial with the ESC Framework and train crowdworkers to be supporters. Only those who pass the examination are admitted to the task.
(2) We require help-seekers to complete a pre-chat survey on their problems and emotions and to provide feedback during and after the conversations.
(3) We devise and use multiple manual or automatic mechanisms to filter out the low-quality conversations after collecting raw dialog data.

\vspace{-1mm}
\subsection{Supporter-specific Tasks}
\vspace{-1mm}
\label{subsec:supporter_task}

\noindent \textbf{Training and Examination} \quad 
To teach crowdworkers how to provide effective emotional support, we designed a tutorial with the ESC Framework. 
Inspired by 7cups (\url{7cups.com}) \cite{baumel2015online}, we developed eleven sub-tasks (3 + 8) to help workers to learn the definitions of the three stages and the eight support strategies.
Each sub-task includes an example conversation excerpt and a corresponding quiz question.
As noted in Section \ref{subsec:framework}, we also informed participants that following a fixed order may not be possible and that they may need to be flexible with adjusting the stage transitions.

\noindent \textbf{Strategy Annotation}\quad
To encourage supporters to use the ESC support strategies during the conversation and to structure the resulting dataset, we ask the supporter to first select a proper strategy that they would like to use according to the dialog context. They are then able to write an utterance reflecting their selected strategy.
We encourage supporters to send multiple messages if they would like to use multiple strategies to provide support.

\noindent\textbf{Post-chat Survey}\quad
After each conversation, the supporter is asked to rate the extent that the seeker goes into detail about their problems on five-point Likert scales.



\vspace{-1mm}
\subsection{Seeker-specific Tasks}
\vspace{-1mm}
\label{subsec:seeker_task}

\noindent\textbf{Pre-chat Survey}\quad
Before each conversation, the help-seeker was asked to complete the following survey:
(1) Problem \& emotion category: the help-seeker should select one problem from 5 options and one emotion from 7 options (the options were based on conversations collected in pilot data collection trials).
(2) Emotion intensity: a score from 1 to 5 (the larger number indicates a more intense emotion).
(3) Situation: open text describing the causes of the emotional problem.
(4) Experience origin: whether the described situation was the current experience of the help-seeker or based on prior life circumstances. 
We found that 75.2\% of conversations originated from the help-seekers' current experiences.

\noindent\textbf{Feedback}\quad
%
During the conversation, the help-seeker was asked to give feedback after every two new utterances they received from the supporter.
Their feedback scored the helpfulness of the supporter messages on a 5-star scale.
We divided each conversation into three phases and calculated the average feedback score for each phase.
The scores in the three phases are 4.03, 4.30, and 4.44 respectively, 
indicating that the supporters were sufficiently trained to effectively help the help-seekers feel better.

\noindent\textbf{Post-chat Survey}\quad
After each conversation, the help-seeker is asked to rate their emotion and the performance of the supporter on the following five-point Likert scales:
(1) Their emotion intensity after the emotional support conversation (a decrease from the intensity before the conversation reflects emotion improvement),
(2) the supporter's empathy and understanding of the help-seeker's experiences and feelings,
and (3) the relevance of the supporter's responses to the conversation topic.

\begin{table}[t]
  \centering
  \scalebox{0.7}{
    \begin{tabular}{p{4.5em}p{16em}p{3.5em}}
        \toprule
        \textbf{Roles} & \textbf{Aspects} & \textbf{Criteria} \\
        \midrule
        \multirow{7}[2]{*}{\tabincell{l}{\textbf{Supporter} \\ $\bm{(\ge 3)}$\textbf{*}}} &
        Understanding the help-seeker's experiences and feelings (\textit{rated by the help-seeker})  & $>=3$ \\
        \cmidrule{2-3} 
        & Relevance of the utterances to the conversation topic (\textit{rated by the help-seeker}) & $>=4$ \\
        \cmidrule{2-3}
        & Average length of utterances  & $>=8$ \\
        \cmidrule{2-3}
        & Improvement in the help-seeker's emotion intensity (\textit{rated by the help-seeker})** & $>=1$ \\
        \midrule
        \multirow{3}[2]{*}{\textbf{Seeker}}
         & Describing details about the own emotional problems (\textit{rated by the supporter}) & not required \\\cmidrule{2-3}
          & Average length of utterances  & $>=6$\\
        \toprule
    \end{tabular}%
  }
\vspace{-1mm}
  \caption{
    Criteria of high-quality conversations.
    * denotes that supporters must meet at least two of the three criteria.
    In **, the improvement of the help-seeker's emotion intensity was calculated by subtracting the intensity after from that before the conversation.
  }
  \label{tab:survey}%
  \vspace{-3mm}
\end{table}%

\vspace{-1mm}
\subsection{Quality Control}
\vspace{-1mm}

We use multiple methods to ensure that the corpus contains high-quality examples of effective emotional support conversations. 

\noindent\textbf{Preliminary Filtering Mechanisms}\quad 
When recruiting participants for the supporter role, we initially received 5,449 applicants, but only 425 (7.8\%) passed the training tutorial.
From the 2,472 conversations that we initially collected, we filtered out those that were not finished by the help-seekers or that had fewer than 16 utterances.
This filtering left 1,342 conversations (54.3\%) for consideration.

\noindent\textbf{Auto-approval Program for Qualified Conversations}\quad 
We carefully designed the auto-approval program, which is the most important part of data quality control. This program uses criteria based on the post-chat survey responses from both roles and the length of utterances, which are summarized in Table \ref{tab:survey}.
These criteria are based on initial human reviewing results.
We show how to choose these auto-approval criteria in Appendix \ref{sec:auto-approval}.
The computed average emotion intensity before conversations is 4.04 and 2.14 after.
Such improvement demonstrates the effectiveness of the emotional support provided by the supporters.
In a small number of conversations, the help-seeker did not finish the post-chat surveys, so we added another criterion for these conversations requiring that the last two feedback scores from the help-seekers are both greater than 4.
Thus, among all the conversations without post-chat surveys, only those who met both (2) and (3) were qualified.
Using these quality criteria, 1,053 (78.5\% of 1,342) of collected conversations were qualified. 

\noindent\textbf{Annotation Correction}\quad 
To further ensure data quality, we reviewed and revised incorrect annotations of support strategy and seeker's emotion intensity.
(1) For strategy annotation correction, we asked new qualified supporters to review and revise annotations on previously collected conversations as necessary, which led to 2,545 utterances (17.1\%) being reviewed.
We manually reviewed annotations where more than 75\% of reviewers disagreed and revised 139 of them. 
(2) According to the auto-approval criteria (Table \ref{tab:criteria}), a conversation can be qualified when the score of the seeker's emotion improvement is less than one, but the other three criteria are satisfied. Upon review, we found this to most often result from seekers mistaking negative emotion intensity as the positiveness of their emotion.
We manually re-checked and revised the emotion intensity of these conversations by using other helpful information, such as the responses to the post-chat survey open question and the seekers' feedback scores during the chat. Of 130 such conversations, 92\% were revised and included in the corpus.

\begin{table}[t]
    \centering
    \scalebox{0.7}{
        \begin{tabular}{lrrr}
        \toprule
        \textbf{Category}                         & \textbf{Total}           & \textbf{Supporter}   & \textbf{Seeker}     \\ \midrule
        \# dialogues                      & 1,053           & - & -    \\
        Avg. Minutes per Chat           & 22.6            & - & -    \\
        \# Workers               & 854             & 425         & 532        \\
        \# Utterances                   & 31,410          & 14,855      & 16,555     \\
        Avg. length of dialogues    & 29.8            & 14.1        & 15.7       \\
        Avg. length of utterances    & 17.8            & 20.2        & 15.7    \\\bottomrule
        \end{tabular}
    }
    \caption{Statistics of ESConv.}
    \label{tab:stats}
  \vspace{-2mm}
\end{table}

\begin{table}[t]
    \centering
      \scalebox{0.7}{
        \begin{tabular}{clrr}
        \toprule
              & \textbf{Categories} & \multicolumn{1}{r}{\textbf{Num}} & \multicolumn{1}{r}{\textbf{Proportion}} \\
        \midrule
        \multirow{6}[2]{*}{\begin{sideways}\textbf{Seeker's Problem}\end{sideways}} & Ongoing Depression & 306   & 29.1\% \\
              & Job Crisis & 233   & 22.1\% \\
              & Breakup with Partner & 216   & 20.5\% \\
              & Problems with Friends & 159   & 15.1\% \\
              & Academic Pressure & 139   & 13.2\% \\
    \cmidrule{2-4}          & Overall & 1,053  & 100.0\% \\
        \midrule
        \multirow{8}[2]{*}{\begin{sideways}\textbf{Seeker's Emotion}\end{sideways}} & Anxiety & 281   & 26.7\% \\
              & Depression & 276   & 26.2\% \\
              & Sadness & 250   & 23.7\% \\
              & Anger & 96    & 9.1\% \\
              & Fear  & 88    & 8.4\% \\
              & Disgust & 32    & 3.0\% \\
              & Shame & 30    & 2.8\% \\
    \cmidrule{2-4}          & Overall & 1,053  & 100.0\% \\
        \midrule
        \multirow{6}[2]{*}{\begin{sideways}\textbf{Seeker's Feedback}\end{sideways}} & 1 (Very Bad) & 71    & 1.1\% \\
              & 2 (Bad) & 183   & 2.9\% \\
              & 3 (Average) & 960   & 15.5\% \\
              & 4 (Good) & 1,855  & 29.9\% \\
              & 5 (Excellent) & 3,144  & 50.6\% \\
    \cmidrule{2-4}          & Overall & 6,213  & 100.0\% \\
        \midrule
        \multirow{9}[2]{*}{\begin{sideways}\textbf{Support Strategy}\end{sideways}} & Question & 3,109  & 20.9\% \\
              & Restatement or Paraphrasing & 883   & 5.9\% \\
              & Reflection of Feelings & 1,156  & 7.8\% \\
              & Self-disclosure   & 1,396  & 9.4\% \\
              & Affirmation and Reassurance & 2,388  & 16.1\% \\
              & Providing Suggestions & 2,323  & 15.6\% \\
              & Information & 904   & 6.1\% \\
              & Others & 2,696  & 18.1\% \\
    \cmidrule{2-4}          & Overall & 14,855  & 100.0\% \\
        \bottomrule
        \end{tabular}%
      }
    \caption{Statistics of all the annotations, including the help-seekers' problems, emotions, feedback, and the support strategies.}
    \label{tab:annotation}
  \vspace{-2mm}
\end{table}

\vspace{-1mm}
\section{Data Characteristics}
\vspace{-1mm}

\subsection{Statistics}
\vspace{-1mm}

The overall statistics of the 1,053 ESConv examples are shown in table \ref{tab:stats}.
Relatively long conversations (avg. 29.8 utterances) indicate that providing effective ES usually requires many turns of interaction and considerably more turns than typical for previous 
emotional chatting \cite{zhou2018emotional} or empathetic dialog \cite{rashkin-etal-2019-towards} datasets.

We also present the statistics of other annotations in Table \ref{tab:annotation}.
Perhaps due to the current outbreak of COVID-19, \textit{ongoing depression} and \textit{job crisis} are the most commonly stated problems for the help-seekers and \textit{depression} and \textit{anxiety} are the most commonly noted emotions. 
From the help-seekers' feedback, we found that they are usually highly satisfied with the emotional support, which further indicates that the training tutorial based on the ESC Framework indeed helps supporters learn to provide effective ES.
We release all these annotations to facilitate further research.

\vspace{-1mm}
\subsection{Strategy Analysis}
\vspace{-1mm}
\label{subsec:strAnalysis}

\noindent\textbf{Lexical Features}\quad 
We extracted lexical features of each strategy by calculating the log odds ratio, informative Dirichlet prior \cite{monroe2008fightin} of all the unigrams and bigrams for each strategy contrasting to all other strategies.
We list the top 5 phrases for each strategy in Figure \ref{fig:framework}. 
Those strategies are all significantly ($z$-score $>$ 3) associated with certain phrases (e.g., \textit{Question} with ``are you'', \textit{Self-disclosure} with ``me'').

\begin{figure}[t]
    \centering
    \includegraphics[width=0.9\linewidth]{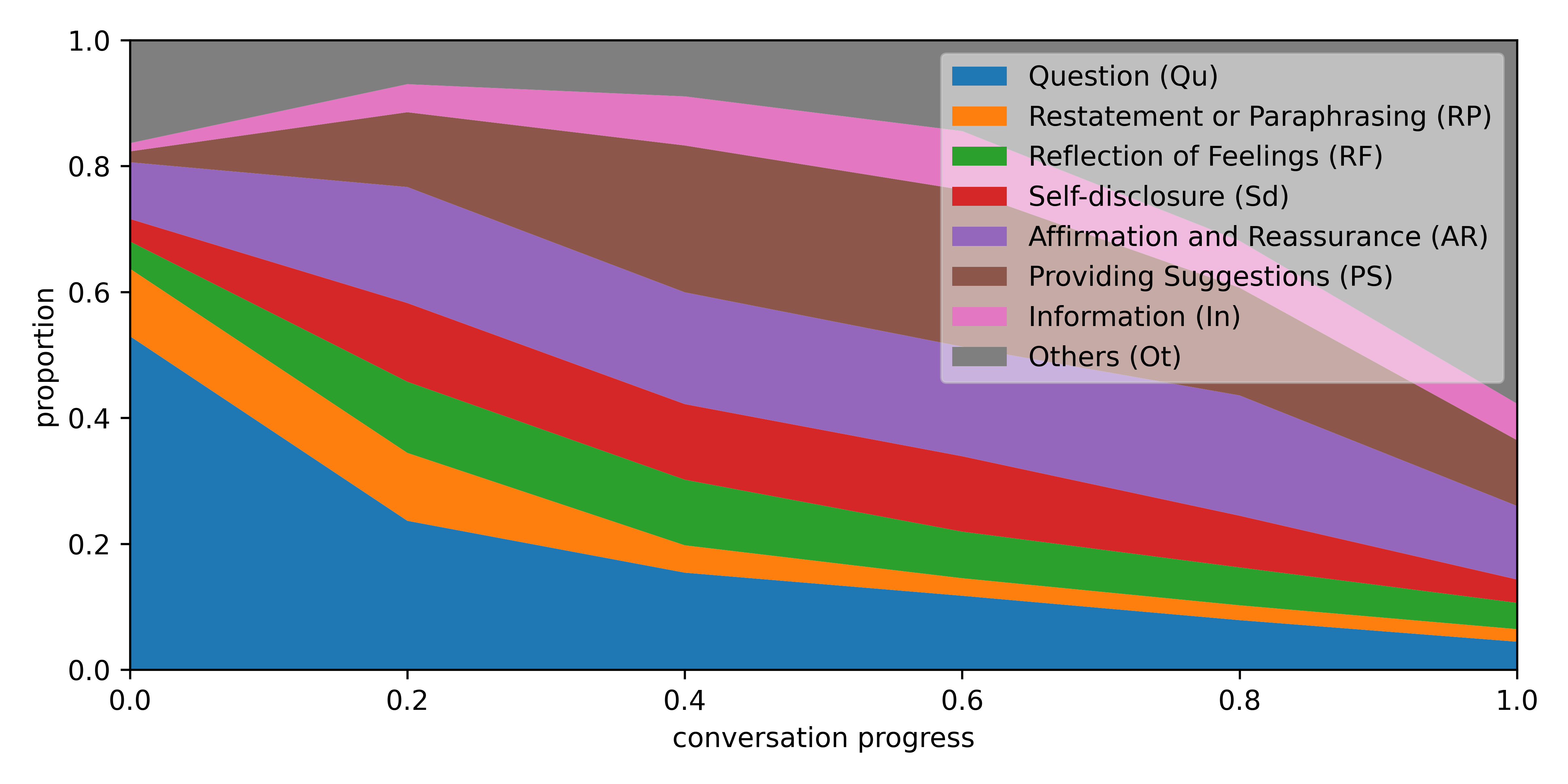}
    \caption{
    The distribution of strategies at different conversation progress. 
    }
    \label{fig:distribution}
    \vspace{-2mm}
\end{figure}

\noindent\textbf{Strategy Distribution}\quad 
We computed the distribution of strategies at different phases of the conversation.
For a conversation with $L$ utterances in total, the $k$-th ($1\le k \le L$) utterance is from the supporter and adopts the strategy $\mathrm{st}$, we say that it locates at the conversation progress $k/L$.
Specifically, we split the conversation progress into six intervals: $[0, 1] = \bigcup_{i=0}^4 [i / 5, (i+1)/5) \bigcup \{ 1 \} $.
Then, for all the conversations in ESConv, we counted the proportions of different strategies in the six intervals.
We split the conversation progress into six intervals: $[0, 1] = \bigcup_{i=0}^4 [i / 5, (i+1)/5) \bigcup \{ 1 \} $ and drew the distributions on the six intervals at six points $i/5 (i=0, \dots,5)$ respectively and connected them, finally obtaining Figure \ref{fig:distribution}.

The supporters generally follow the stage order suggested by the ESC Framework (Figure \ref{fig:framework}), but there is also flexible adjustment of stages and adoption of strategies.
For instance, at the early phase of conversation, the supporters usually adopt exploratory strategies such as \textit{Question}.
After knowing help-seekers' situations, the supporters tend to provide their opinions (such as \textit{Providing Suggestions}).
Throughout the entire conversation, the comforting strategies (such as \textit{Affirmation and Reassurance}) are used and label a relatively constant proportion of messages.

\noindent\textbf{Strategy Transition}\quad 
We present the top-5 most frequent strategy transitions with 3 / 4 hops in Appendix (Table \ref{tab:transition}).
These transitions indicate that, as the tutorial of ESC framework trains, supporters usually ask questions and explore the help-seekers' situations before comforting the help-seekers.

\vspace{-1mm}
\section{Experiments}
\vspace{-1mm}

Our experiments focus on two key questions:
(1) How much can ESConv with strategy annotation improve state-of-the-art generative dialog models?
(2) Can these models learn to provide effective emotional support from ESConv?

\vspace{-1mm}
\subsection{Backbone Models}
\vspace{-1mm}

We used two state-of-the-art pre-trained models as the backbones of the compared variant models:

\noindent\textbf{BlenderBot}\quad 
BlenderBot \cite{roller2020recipes} is an open-domain conversational agent trained with multiple communication skills, including empathetic responding. 
As such, BlenderBot should be capable of providing ES for users to some extent. 
We used the small version\footnote{\url{https://huggingface.co/facebook/BlenderBotbot_small-90M}} of BlenderBot in experiments, because the larger versions have the limitation of maximum context length 128, which we found harms the model performance and response coherence.

\noindent\textbf{DialoGPT}\quad 
We additionally evaluated DialoGPT \cite{zhang-etal-2020-dialogpt}, which is a GPT-2-based model pre-trained on large-scale dialog corpora.
We used the small version\footnote{\url{https://huggingface.co/microsoft/DialoGPT-small}}.

\vspace{-1mm}
\subsection{Variant Models}
\vspace{-1mm}
\label{subsec:variant}

Taking each of the above pre-trained models as the backbone, we built the following variant models:

\noindent\textbf{Vanilla}\quad 
Directly fine-tuning the backbone model on ESConv with no access to strategy annotations.
Formally, suppose the flattened dialog history is $\mathbf{x}$ and the response to be generated is $\mathbf{y}$, we maximize the conditional probability: $\mathbb{P}(\mathbf{y}|\mathbf{x}) = \prod_{i=1}^{|\mathbf{y}|} \mathbb{P} \left( y_i | \mathbf{x}, \mathbf{y}_{\le i} \right)$.

\noindent\textbf{Variants with strategy}\quad 
To incorporate the strategy annotation into the backbone model, we used a special token to represent each strategy.
For each utterance $\mathbf{y}$ from the supporters, we appended the corresponding strategy token before this utterance: $\tilde{\mathbf{y}}=\mathrm{[st]} \oplus \mathbf{y}$, where $\mathrm{[st]}$ denotes the special token of the used strategy.
Then, taking the flattened dialog history $\mathbf{x}$ as input, the model generates the response conditioned on the first predicted (or designated) strategy token: $\mathbb{P}(\tilde{\mathbf{y}}|\mathbf{x}) = \mathbb{P}(\mathrm{[st]} |\mathbf{x}) \prod_{i=1}^{|\mathbf{y}|} \mathbb{P} \left( y_i | \mathbf{x}, \mathrm{[st]}, \mathbf{y}_{< i} \right)$.

We studied three variants that use strategy annotation in the later experiments.
(1) \textbf{Oracle}: responses are generated conditioned on the gold reference strategy tokens. 
(2) \textbf{Joint}: responses are generated conditioned on predicted (sampled) strategy tokens. 
(3) \textbf{Random}: responses are generated conditioned on randomly selected strategies. 
Implementation details are in Appendix \ref{sec:implementation}.

\begin{table}[t]
  \centering
  \scalebox{0.7}{
    \begin{tabular}{lccccc}
    \toprule
    \textbf{Backbones} & \textbf{Variants} & \textbf{PPL} & \textbf{B-2} & \textbf{R-L} & \textbf{Extrema} \\
    \midrule
    \multirow{3}[3]{*}{\textbf{DialoGPT}} & Vanilla & 15.51  & 5.13  & 15.26  & 49.80  \\
\cmidrule{2-6}          & Joint & -     & 5.00  & 15.09  & 49.97  \\
\cmidrule{2-6}          & Oracle & 15.19  & 5.52  & 15.82  & 50.18  \\
    \midrule
    \multirow{3}[3]{*}{\textbf{BlenderBot}} & Vanilla & 16.23  & 5.45  & 15.43  & 50.49  \\
\cmidrule{2-6}          & Joint & -     & 5.35  & 15.46  & 50.27  \\
\cmidrule{2-6}          & Oracle & 16.03  & \textbf{6.31} & \textbf{17.90} & \textbf{51.65} \\
    \bottomrule
    \end{tabular}%
  }
\vspace{-1mm}
  \caption{
  Results of automatic evaluation.
  The results in \textbf{bold} are significantly better than all the competitors (Student's t-test, $p$-value $<$ 0.05).
  }
  \label{tab:automatic}
    \vspace{-3mm}
\end{table}

\vspace{-1mm}
\subsection{Automatic Evaluation}
\vspace{-1mm}

To investigate the impact of utilizing support strategies on the model performance with either BlenderBot or DialoGPT as the backbone, we compared the performance of the Vanilla, Joint, and Oracle variants described above.
The automatic metrics we adopted include perplexity (\textbf{PPL}), BLEU-2 (\textbf{B-2}) \cite{papineni-etal-2002-bleu}, ROUGE-L (\textbf{R-L}) \cite{lin-2004-rouge},
and the BOW Embedding-based \cite{liu-etal-2016-evaluate} \textbf{Extrema} matching score.
The metrics except PPL were calculated with an NLG evaluation toolkit\footnote{\url{https://github.com/Maluuba/nlg-eval}} \cite{sharma2017relevance} with responses tokenized by NLTK\footnote{\url{https://www.nltk.org/}} \cite{loper2002nltk}.

There are three major findings from the experiments (Table \ref{tab:automatic}).
(1) The Oracle models are significantly superior to the Vanilla models on all the metrics, indicating the great utility of support strategies.
(2) The Joint models obtain sightly lower scores than the Vanilla models, as, if the predicted strategy is different from the ground truth, the generated response will be much different from the reference response.
However, learning to predict strategies is important when there are no ground truth labels provided, and we will further investigate the performance of the Joint model in human interactive evaluation (Section \ref{subsec:interactive}).
(3) The BlenderBot variants consistently perform better than the DialoGPT ones, indicating that BlenderBot is more suitable for the ESC task. 
Thus, in the subsequent human evaluation, we will focus evaluation on the BlenderBot variants.

\vspace{-1mm}
\subsection{Human Interactive Evaluation}
\vspace{-1mm}
\label{subsec:interactive}

We recruited participants from Amazon Mechanical Turk to chat with the models.
The online tests were conducted on the same platform as our data collection, but with the role of supporter taken by a model.
Each participant chatted with two different models that were randomly ordered to avoid exposure bias.
Participants were asked to compare the two models based on the following questions:
(1) \textbf{Fluency}: which bot's responses were more fluent and understandable?
(2) \textbf{Identification}: which bot explored your situation more in depth and was more helpful in identifying your problems?
(3) \textbf{Comforting}: which bot was more skillful in comforting you?
(4) \textbf{Suggestion}: which bot gave you more helpful suggestions for your problems?
(5) \textbf{Overall}: generally, which bot's emotional support do you prefer?
The metrics in (2), (3), and (4) correspond to the three stages in the ESC Framework.

We compare three pairs of models:
(a) Joint vs. BlenderBot (without fine-tuning on ESConv), 
(b) Joint vs. Vanilla, 
and (c) Joint vs. Random (using randomly selected strategies).
To better simulate the real strategy occurrence, the Random model randomly selects a strategy following the strategy distribution in ESConv (Table \ref{tab:annotation}).

\begin{table}[t]
    \centering
    \scalebox{0.7}{
    \begin{tabular}{lllllll}
        \toprule
        \multirow{2}[0]{*}{\textbf{Joint vs.}} & \multicolumn{2}{c}{\textbf{w/o ft}} & \multicolumn{2}{c}{\textbf{Vanilla}} & \multicolumn{2}{c}{\textbf{Random}} \\
              & \textbf{Win} & \textbf{Lose} & \textbf{Win} & \textbf{Lose} & \textbf{Win} & \textbf{Lose} \\
        \midrule
        \textbf{Fluency} & \boldmath{}\textbf{71$^\ddag$}\unboldmath{} & 24    & \boldmath{}\textbf{52$^\dag$}\unboldmath{} & 35    & \boldmath{}\textbf{53$^\dag$}\unboldmath{} & 35 \\
        \textbf{Identification} & \boldmath{}\textbf{65$^\ddag$}\unboldmath{} & 25    & \textbf{50} & 34    & \boldmath{}\textbf{54$^\dag$}\unboldmath{} & 37 \\
        \textbf{Comforting} & \boldmath{}\textbf{75$^\ddag$}\unboldmath{} & 20    & \boldmath{}\textbf{54$^\ddag$}\unboldmath{} & 34    & \textbf{47} & 39 \\
        \textbf{Suggestion} & \boldmath{}\textbf{72$^\ddag$}\unboldmath{} & 21    & \textbf{47} & 39    & \boldmath{}\textbf{48$^\dag$}\unboldmath{} & 27 \\
        \midrule
        \textbf{Overall} & \boldmath{}\textbf{73$^\ddag$}\unboldmath{} & 20    & \boldmath{}\textbf{51$^\dag$}\unboldmath{} & 34    & \boldmath{}\textbf{56$^\ddag$}\unboldmath{} & 36 \\
        \bottomrule
    \end{tabular}%
    }
\vspace{-1mm}
    \caption{
    Results of the human interactive evaluation.
    Ties are not shown.
    All the models use BlenderBot as the backbone.
    `w/o ft' denotes the BlenderBot model without fine-tuning on ESConv.
    The Joint model outperforms all the competitors on all the metrics (sign test, ${\dag} / {\ddag}$ denote $p$-value $< 0.1 / 0.05$ respectively).}
    \label{tab:interactive}
    \vspace{-3mm}
\end{table}

Each pair of models was compared by 100 conversations with human participants (Table \ref{tab:interactive}). 
The results of comparison (a) show that BlenderBot's capability of providing ES is significantly improved on all the metrics after being fine-tuned on ESConv.
From comparison (b), we found that utilizing strategies can better comfort the users.
The results of comparison (c) also demonstrate that the proper timing of strategies is critical to help users identify their problems and to provide effective suggestions.
In general, through being fine-tuned with the supervision of strategy prediction on ESConv, the pre-trained models become preferred by the users, which proves the high-quality and utility of ESConv.

\vspace{-1mm}
\subsection{Further Analysis of Human Interactive Evaluation}
\vspace{-1mm}
\label{subsec:analysis}

\begin{figure}[t]
    \centering
    \includegraphics[width=0.9\linewidth]{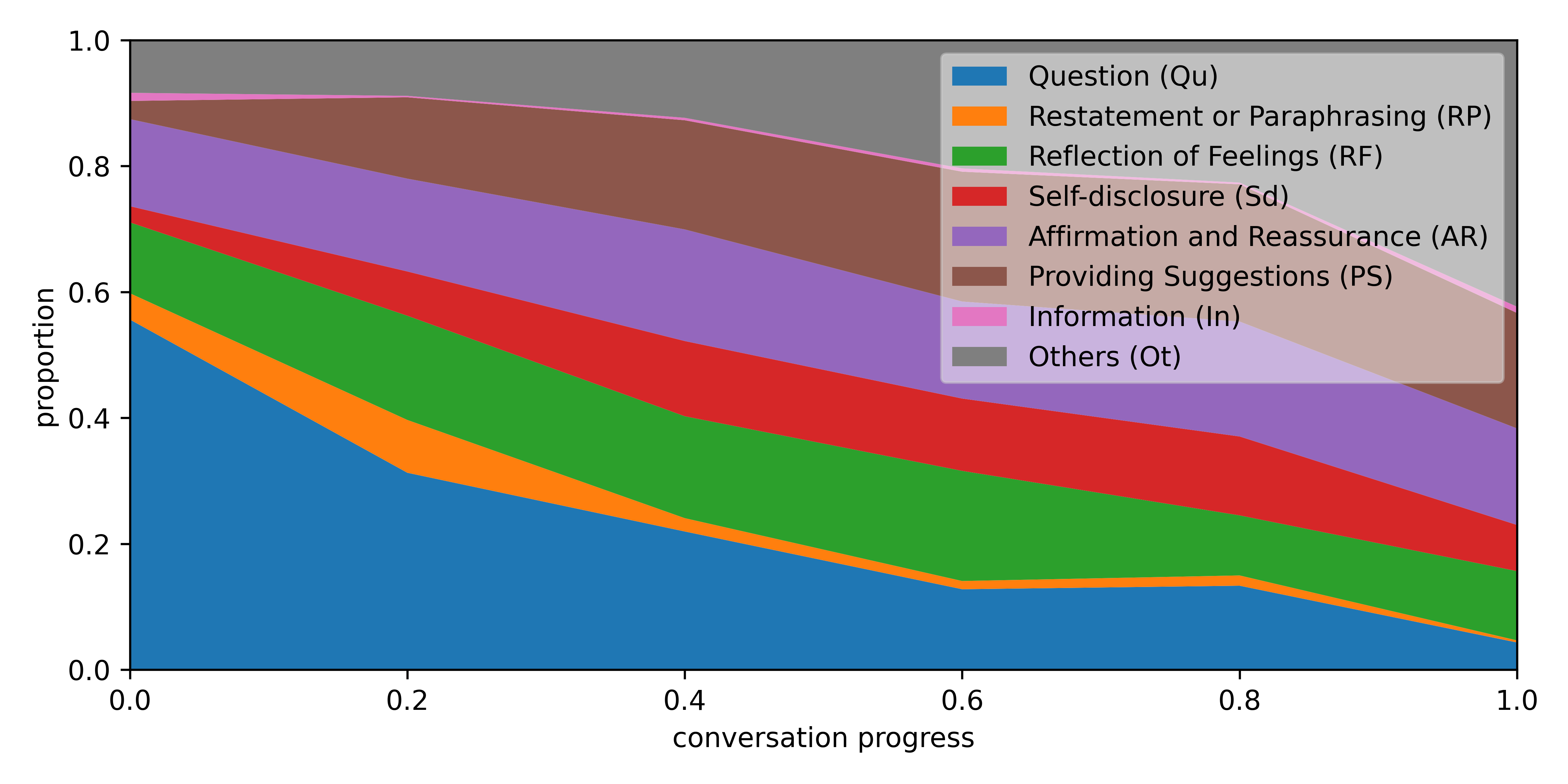}
\vspace{-1mm}
    \caption{
    The Joint model's generation distribution.
    The meanings of all the graphics and abbreviations are consistent with Figure \ref{fig:distribution}.}
    \label{fig:botdistribution}
    \vspace{-3mm}
\end{figure}

In this section, we explore what the dialog models learned from ESConv. 
\textbf{Firstly}, we analyzed the strategy distribution based on the 300 dialogs between users and the Joint model in human interactive experiments.
We can see in Figure \ref{fig:botdistribution} (the calculation was consistent with Figure \ref{fig:distribution}), the strategies that the Joint model adopted have a very similar distribution compared with the truth distribution in ESConv (Figure \ref{fig:distribution}). 
It provides important evidence that models mimic strategy selection and utilization as human supporters do to achieve more effective ES.
\textbf{Secondly}, we present a case study in Figure \ref{fig:case}.
We see in cases that the Joint model provides more supportive responses and uses more skills in conversation, while BlenderBot without fine-tuning seems not to understand the user's distress very well and prefers to talk more about itself. 
This may imply that having more supportive responses and a diverse set of support strategies are crucial to effective emotional support.

\vspace{-1mm}
\section{Conclusion}
\vspace{-1mm}

In this work, we define the task of Emotional Support Conversation and present an ESC Framework.
The ESC Framework is adapted from the Helping Skills Theory into a dialog system setting, which characterizes three stages with corresponding support strategies useful at each stage.
We then construct an Emotional Support Conversation dataset, ESConv.
We carefully design the process of data collection and devise multiple mechanisms to ensure the effectiveness of ES in conversations.
Finally, we evaluate the ES ability with state-of-the-art dialog models.
Experimental results show the potential utility of ESConv in terms of improving dialog systems' ability to provide effective ES.
Our work can facilitate future research of ES dialog systems, as well as improve models for other conversation scenarios where emotional support plays an important role. Strategy selection and realization, user state modeling, and task evaluation are important directions for further research.

\vspace{-1mm}
\section*{Acknowledgments}
\vspace{-1mm}

This work was supported by the NSFC projects (Key project with No. 61936010 and regular project with No. 61876096). 
This work was also supported by the Guoqiang Institute of Tsinghua University, with Grant No. 2019GQG1 and 2020GQG0005.

\vspace{-1mm}
\section*{Ethical Considerations}
\vspace{-1mm}


There are many types and levels of support that humans can seek to provide, e.g., professional versus peer support, and some of these levels may be inappropriate, unrealistic, and too risky for systems to deliver. However, as dialog systems become more common in daily use, opportunities will arise when at least some basic level of supportive statements may be required. In developing the ESC Framework, we have carefully considered which elements of conversational support may be relevant for a dialog system and omitted elements that are clear oversteps. Considerable additional work is needed to determine what are appropriate levels of support for systems to provide or that can be expected from systems, but our work provides a cautious, yet concrete, step towards developing systems capable of reasonably modest levels of support. The corpus we construct can also provide examples to enable future work that probes the ethical extent to which systems can or should provide support. In addition to these broader ethical considerations, we have sought to ethically conduct this study, including by transparently communicating with crowdworkers about data use and study intent, compensating workers at a reasonable hourly wage, and obtaining study approval from the Institutional Review Board. 


\bibliographystyle{acl_natbib}
\bibliography{anthology,acl}

\newpage

\appendix

\section{Data Example from ESConv}
\label{app:example}
\begin{figure}[h]
    \centering
    \includegraphics[width=\linewidth]{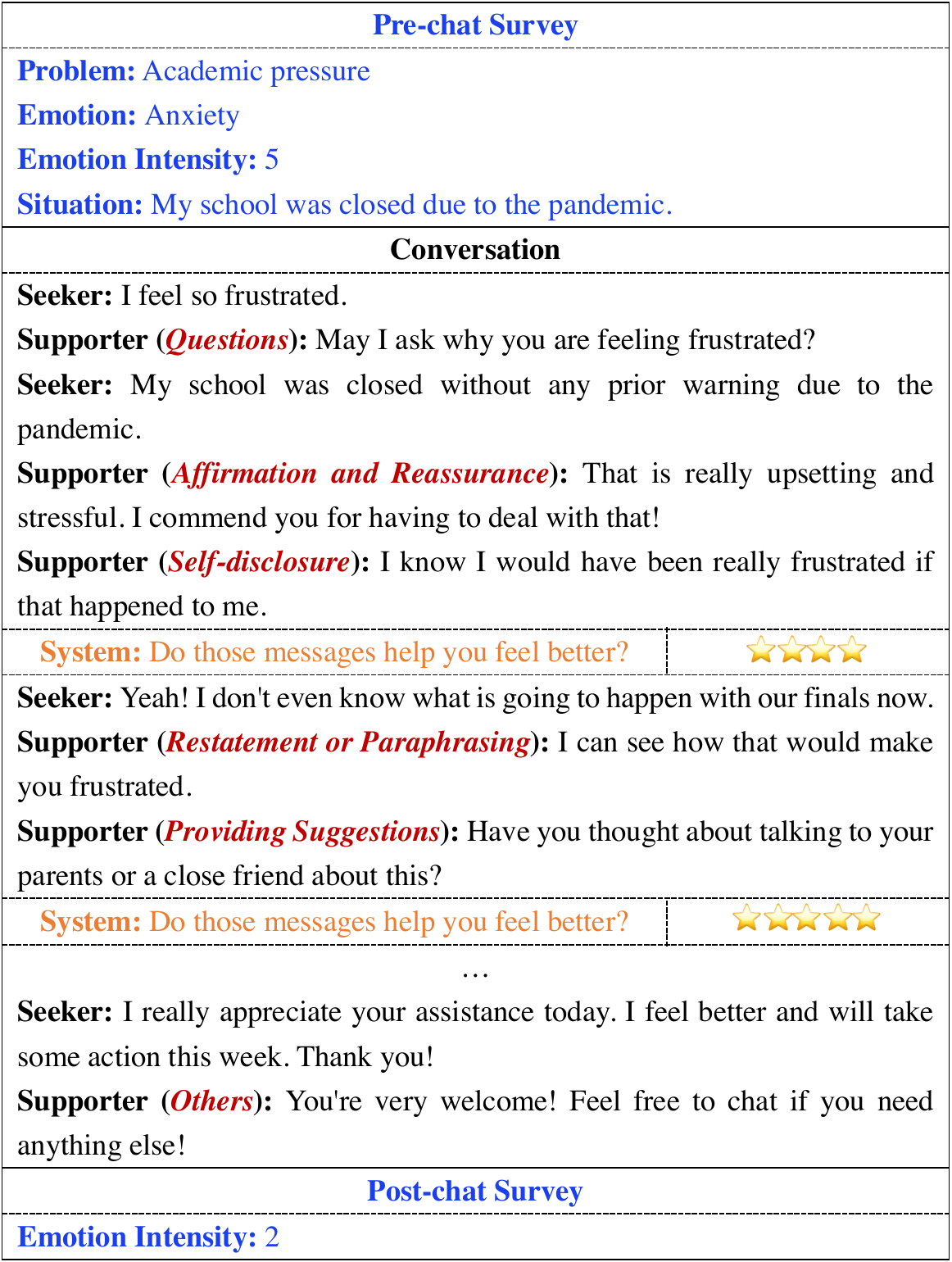}
    \caption{Data example from ESConv. 
    {\color{blue} \textbf{Blue text}}: the help-seeker's pre-chat survey.
    \textcolor[RGB]{190,37,14}{\textbf{Red text}}: strategies used by the supporter.
    {\color{orange} \textbf{Orange text}}: the question that the systems ask help-seeker to evaluate the helpfulness per two utterances from the supporter. Thus the stars denote the seeker's feedback score.}
    \label{fig:example}
\end{figure}

Here we detail the conversation that Figure \ref{fig:intro_example} demonstrates to show the annotations that our dataset contains. The detailed example can be seen in Figure \ref{fig:example}. Each pre-chat survey of conversation is labeled its problem category, emotion category, emotion intensity, and a brief of the situation of the seeker. In the context of each conversation, the strategies used by supporters are labeled and the seeker's feedback score per two utterances of the supporter's responses are also given in our dataset. Note that not all conversations have the label of emotion intensity after the conversation. It is because some seekers don't finish the post-chat survey but we still include such conversations into our dataset due to their high quality that meets our criteria.

\begin{table}[t]
  \centering
  \scalebox{0.7}{
    \begin{tabular}{clr}
        \toprule
              & \textbf{Strategy Transition} & \textbf{Proportion} \\
        \midrule
        \multirow{5}[0]{*}{\textbf{3-Hop}} & Qu $\to$ AR $\to$ Qu & 19.65 ‰ \\
              & Qu $\to$ RP $\to$ Qu & 14.55 ‰ \\
              & Qu $\to$ RP $\to$ AR & 12.37 ‰ \\
              & AR $\to$ Qu $\to$ AR & 11.96 ‰ \\
              & Ot $\to$ Qu $\to$ RP & 11.64 ‰ \\
        \midrule
        \multirow{5}[0]{*}{\textbf{4-Hop}} & Qu $\to$ AR $\to$ Qu $\to$ AR & 7.00 ‰ \\
              & AR $\to$ Qu $\to$ AR $\to$ Qu & 5.13 ‰ \\
              & Ot $\to$ Qu $\to$ RP $\to$ Qu & 4.20 ‰ \\
              & PS $\to$ Ot $\to$ PS $\to$ Ot & 3.85 ‰ \\
              & Qu $\to$ RP $\to$ AR $\to$ Qu & 3.85 ‰ \\
        \bottomrule
    \end{tabular}%
  }
  \caption{
  Proportions of top-5 strategy transitions in supporter utterances. Abbreviations are consistent with Figure \ref{fig:distribution}.
  }
  \label{tab:transition}%
    \vspace{-2mm}
\end{table}%

\section{Definitions of Strategies}
\label{sec:definition}

\noindent \textbf{Question}\quad 
Asking for information related to the problem to help the help-seeker articulate the issues that they face. Open-ended questions are best, and closed questions can be used to get specific information.

\noindent \textbf{Restatement or Paraphrasing}\quad 
A simple, more concise rephrasing of the help-seeker’s statements that could help them see their situation more clearly.

\noindent \textbf{Reflection of
Feelings
}\quad 
Articulate and describe the help-seeker’s feelings.

\noindent \textbf{Self-disclosure}\quad 
Divulge similar experiences that you have had or emotions that you share with the help-seeker to express your empathy.

\noindent \textbf{Affirmation and Reassurance}\quad 
Affirm the help-seeker’s strengths, motivation, and capabilities and provide reassurance and encouragement.

\noindent \textbf{Providing Suggestions}\quad 
Provide suggestions about how to change, but be careful to not overstep and tell them what to do.

\noindent \textbf{Information}\quad 
Provide useful information to the help-seeker, for example with data, facts, opinions, resources, or by answering questions.

\noindent \textbf{Others}\quad 
Exchange pleasantries and use other support strategies that do not fall into the above categories.

\section{Implementation Details}
\label{sec:implementation}

The implementation of all models was based on Transformer library\footnote{\url{https://github.com/huggingface/transformers}} \cite{wolf-etal-2020-transformers}. 
We split ESConv into the sets of training / validation / test with the proportions of 6:2:2.
since the conversations in ESConv usually have long turns, we cut each dialog into conversation pieces with 5 utterances, which contain one supporter's response and the preceding 4 utterances.
During training, we trained all the models with Adam \cite{kingma2014adam} optimizer with learning rate $5e^{-5}$.
All the models were trained for 5 epochs, and the checkpoints with the lowest perplexity scores on the validation set were selected for evaluation. 
During inference, we masked other tokens and sampled a strategy token at the first position of the response.
For the Random variant models, we sampled strategies randomly following the strategy distribution in ESConv, which is reported in Table \ref{tab:annotation}.
The response were decoded by Top-$k$ and Top-$p$ sampling with $p=0.9$ \cite{holtzman2019curious}, $k=30$, temperature $\tau=0.7$, and the repetition penalty 1.03.

\begin{table*}[htbp]
  \centering
  \resizebox{0.8\textwidth}{!}{
    \begin{tabular}{llllllllll}\toprule
    \multicolumn{6}{c}{\textbf{Auto-approval Rule}}       & \multicolumn{4}{c}{\textbf{Consistency}} \\\cmidrule(lr){1-6}\cmidrule(lr){7-10}
    \multicolumn{4}{c}{Supporter} & \multicolumn{2}{c}{Seeker} & \multicolumn{1}{c}{\multirow{2}[0]{*}{Human1}} & \multicolumn{1}{c}{\multirow{2}[0]{*}{Human2}} & \multicolumn{1}{c}{\multirow{2}[0]{*}{Human3}} & \multicolumn{1}{c}{\multirow{2}[0]{*}{Average}} \\\cmidrule(lr){1-4}\cmidrule(lr){5-6}
    \multicolumn{1}{l}{Improvement} & \multicolumn{1}{l}{Avg. Length} & \multicolumn{1}{l}{Empathy} & \multicolumn{1}{l}{Relevance} & Detail & \multicolumn{1}{l}{Avg. Length} &       &       &       &  \\\midrule
    \textbf{1} & \textbf{8} & \textbf{3} & \textbf{4} & \textbf{-} & \textbf{6} & \textbf{0.545} & \textbf{0.659} & \textbf{0.525} & \textbf{0.576 } \\
    2     & 8     & 3     & 4     & -     & 6     & 0.505 & 0.566 & 0.486 & 0.519  \\
    1     & 8     & 4     & 4     & -     & 6     & 0.539 & 0.602 & 0.519 & 0.553  \\
    1     & 8     & 2     & 4     & -     & 6     & 0.539 & 0.618 & 0.570  & 0.576  \\
    1     & 8     & 3     & 3     & -     & 6     & 0.546 & 0.630  & 0.526 & 0.567  \\
    1     & 8     & 3     & 5     & -     & 6     & 0.575 & 0.640  & 0.555 & 0.590  \\
    1     & 8     & 3     & 4     & -     & 7     & 0.539 & 0.602 & 0.473 & 0.538  \\
    1     & 8     & 3     & 4     & -     & 5     & 0.520  & 0.551 & 0.501 & 0.524  \\
    1     & 8     & 3     & 4     & \multicolumn{1}{l}{3} & 6     & 0.505 & 0.653 & 0.531 & 0.563  \\
    1     & 8     & 3     & 4     & \multicolumn{1}{l}{2} & 6     & 0.527 & 0.640  & 0.508 & 0.558  \\
    1     & 8     & 3     & 4     & \multicolumn{1}{l}{4} & 6     & 0.457 & 0.599 & 0.482 & 0.513  \\
    1     & 9     & 3     & 4     & -     & 6     & 0.510  & 0.621 & 0.490  & 0.540  \\
    1     & 7     & 3     & 4     & -     & 6     & 0.515 & 0.633 & 0.495 & 0.548  \\\bottomrule
    \end{tabular}%
  }
  \caption{The agreement score between each ``rule'' annotator and three human annotators. 
  The aspects are consistent with Table \ref{tab:survey}. The first rule means: 1) The supporter improves the help-seeker's emotional state as least one score (emotional improvement is calculated as the help-seeker's emotion intensity before the conversation minus the emotion intensity after the conversation). 2) The average length of the supporter's utterances is no less than eight. 3) The empathy score of the supporter's performance is no less than three. 4) The topic relevance score of the supporter's performance is no less than four. 5) The score that measures how well the help-seeker goes into detail about their problems is not required to be used in the rule (denoted as `-'). (6) The average length of the help-seeker's utterances is no less than six.}
  \label{tab:criteria}%
\end{table*}%

\begin{figure*}[t]
    \centering
    \includegraphics[width=1\linewidth]{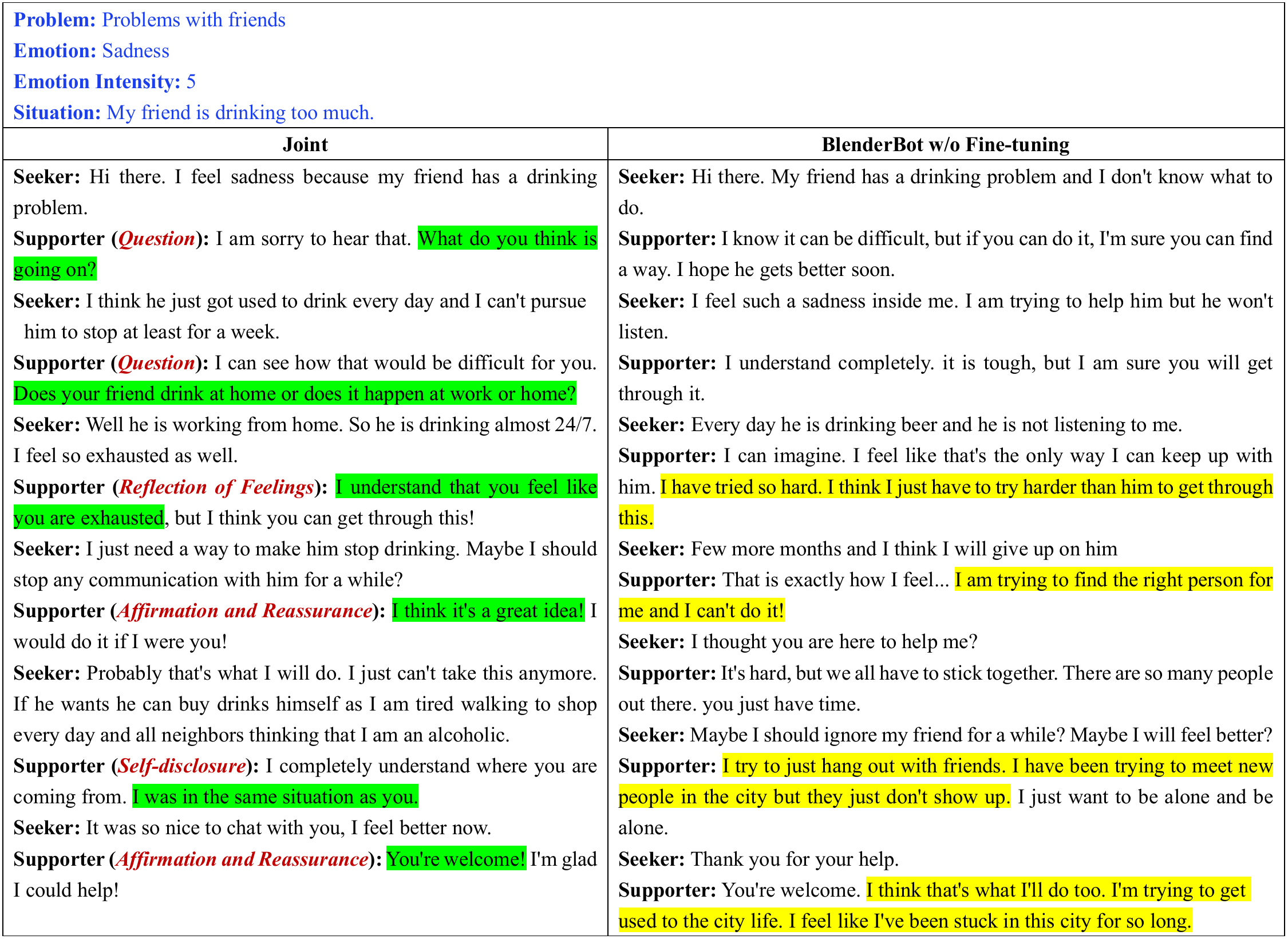}
    \caption{Case Study. 
    {\color{blue} \textbf{Blue text}}: the help-seeker's pre-chat survey.
    \textcolor[RGB]{190,37,14}{\textbf{Red text}}: strategies predicted by the model.
    \colorbox{green}{\textbf{Green background}}: utterances that corresponding to the used strategies.
    \colorbox{yellow}{\textbf{Yellow background}}: utterances where the model talks about itself and is not user-centered. 
    }
    \label{fig:case}
    \vspace{-2mm}
\end{figure*}

\section{Auto-Approval Criteria}
\label{sec:auto-approval}

\begin{figure*}[t]
    \centering
    \subfloat[]{
    \begin{minipage}[l]{0.49\linewidth}
    \includegraphics[width=1\linewidth]{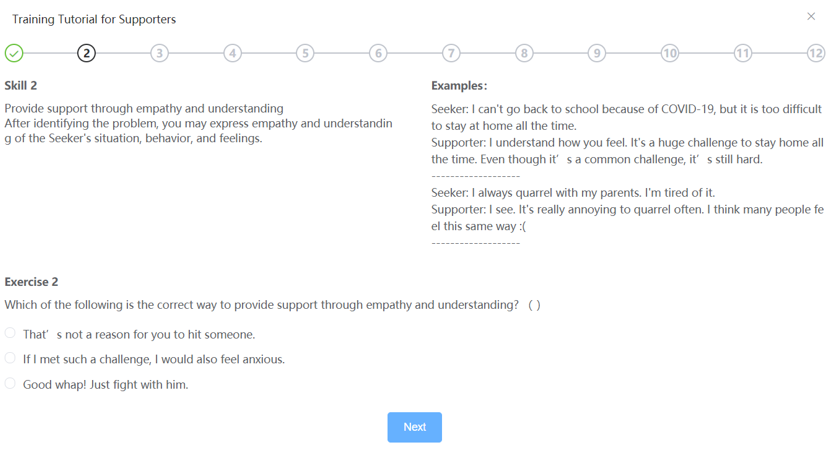}
    \end{minipage}}
    \subfloat[]{
    \begin{minipage}[l]{0.49\linewidth}
    \includegraphics[width=1\linewidth]{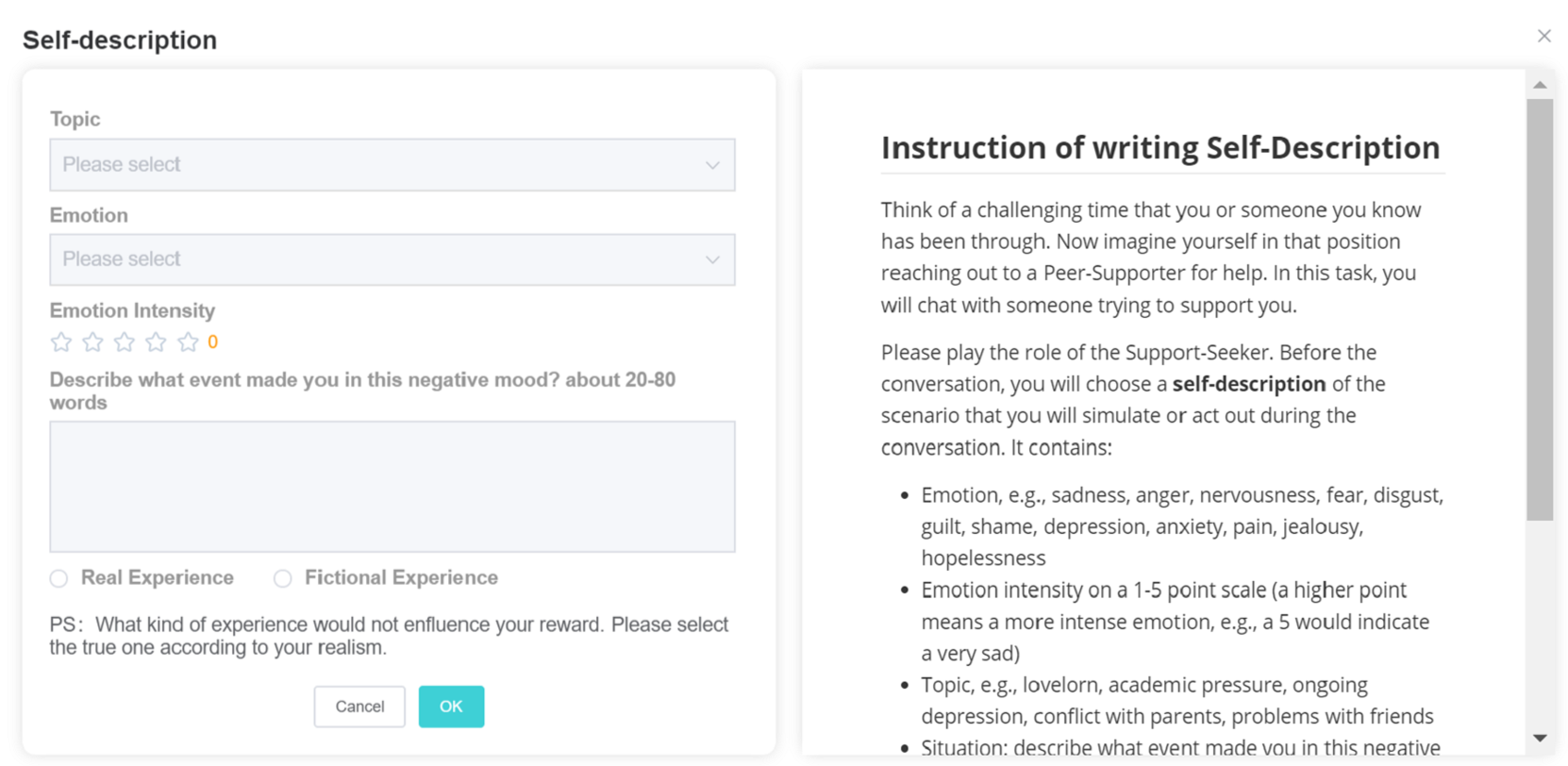}
    \end{minipage}
    }
    \\
    \subfloat[]{
    \begin{minipage}[l]{0.49\linewidth}
    \includegraphics[width=1\linewidth]{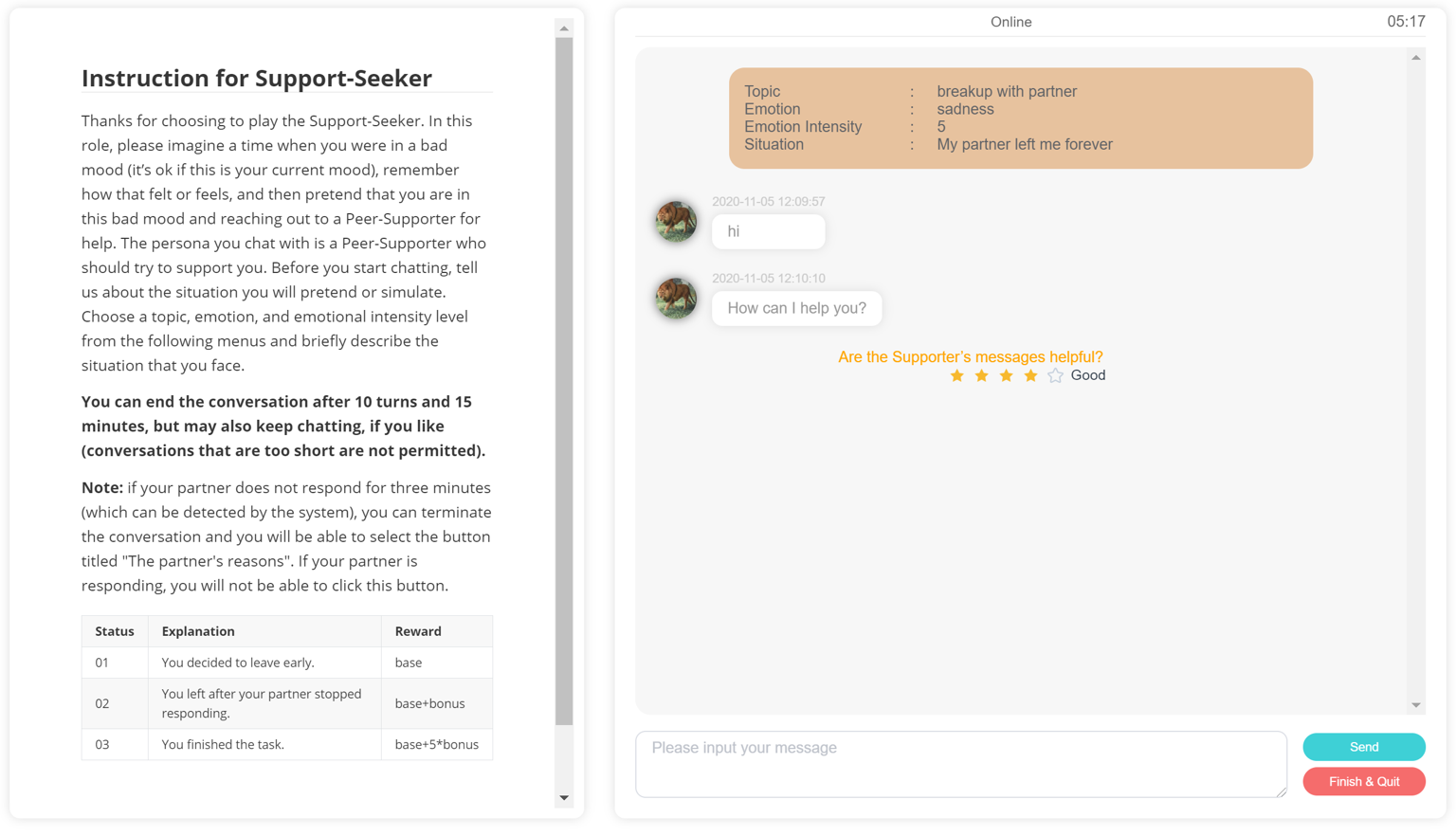}
    \end{minipage}}
    \subfloat[]{
    \begin{minipage}[l]{0.49\linewidth}
    \includegraphics[width=1\linewidth]{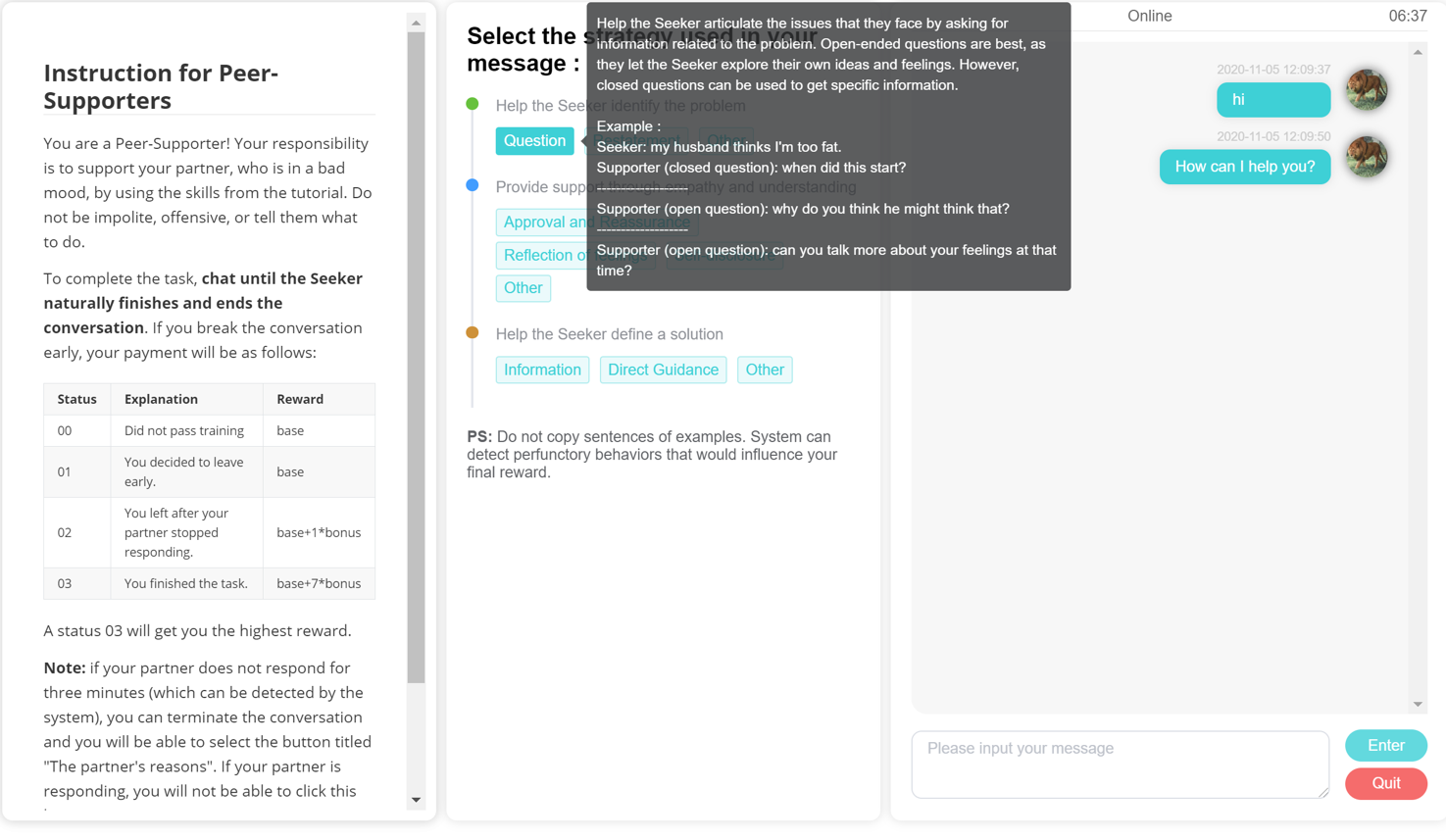}
    \end{minipage}
    }
    \\
   \subfloat[]{
    \begin{minipage}[l]{0.49\linewidth}
    \includegraphics[width=1\linewidth]{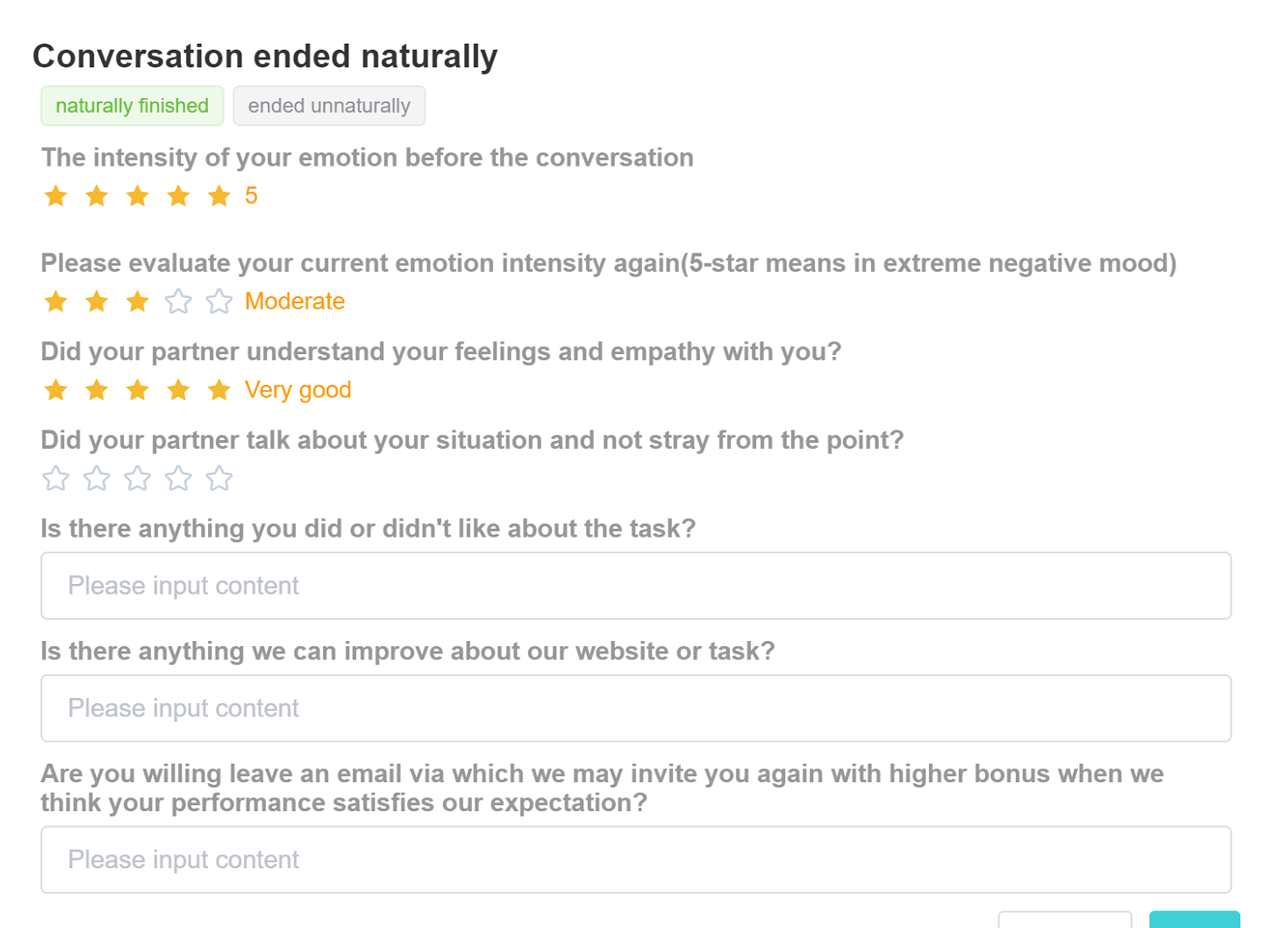}
    \end{minipage}
    }
    \caption{
    (a) Support strategy training. 
    (b) Pre-chat survey. 
    (c) The help-seeker's chatting interface. 
    (d) The supporter's chatting interface. 
    (e) Post-chat survey.
    }
    \label{fig:surface}
    \vspace{-2mm}
\end{figure*}

To establish each criterion of the auto-approval program as shown in the main paper (Section 3.4), we searched the most suitable thresholds for each filtering rule.
We recruited three well-trained human annotators, who have also received the same training procedures as the supporter applicants did.
We then randomly sampled 100 conversations from our dataset and asked the three annotators to judge whether the conversations are qualified for providing effective emotional support.
Next, we utilized the post-survey results and the lengths of speaker utterances to choose suitable thresholds for filtering rules. We then treated each auto-filtering rule as a rule annotator and computed the Cohen's Kappa \cite{cohen1960coefficient} score between the rule annotator and each human annotator.

The agreement scores in Table \ref{tab:criteria} are Cohen's Kappa consistency among the agreement scores between each rule annotator and the three human annotators.
We selected the thresholds that lead to the second-highest agreement score with human annotators and used these thresholds in the filtering rules.
We didn't use the set of thresholds that has the highest agreement score because the rule based on these thresholds is stricter so that many conversations would be filtered out. However, the second-highest score is only slightly lower than the highest so the rule based on the thresholds of second-highest score can remain more qualified conversations with little accepted cost.
As a result, a qualified conversation requires that the supporter must meet at least three of all the four criteria, and the help-seeker must satisfy both of the two corresponding criteria.
The final 'rule' annotator combines the two conditions, and the averaged agreement score between the final rule annotator and the three human annotators is 0.576, indicating significant agreement.

\section{Interface of Data Collection Platform}
\label{app:surface}

To facilitate readers to have an intuitive understanding of our data collection process, we present an interface diagram of some important steps in the data collection process in Figure \ref{fig:surface}, which contains the surfaces of support strategy training, supporter's chatting, help-seeker's pre-chat survey, help-seeker's chatting, and post-survey.

\end{document}